\documentclass{bmvc2k}

\title{H-NeXt: The next step towards roto-translation invariant networks}

\addauthor{Tomáš Karella}{karella@utia.cas.cz}{1}
\addauthor{Filip Šroubek}{sroubekf@utia.cas.cz}{1}
\addauthor{Jan Flusser}{flusser@utia.cas.cz}{1}
\addauthor{Jan Blažek}{blazek@utia.cas.cz}{1}
\addauthor{Vašek Košík}{kosik@utia.cas.cz}{1}

\addinstitution{
Institute of Information Theory and Automation \\ 
Czech Academy of Sciences \\ 
Pod Vodárenskou věží 4 \\
Prague 8, Czech Republic
}

\runninghead{Karella, Šroubek, Flusser, Blažek, Košík}{H-NeXt}

\def\eg{\emph{e.g}\bmvaOneDot}
\def\Eg{\emph{E.g}\bmvaOneDot}
\def\etal{\emph{et al}\bmvaOneDot}
\usepackage[nolist]{acronym}
\usepackage{todonotes}
\usepackage{amsmath,amsthm,mathtools, bbm,amssymb}
\usepackage{makecell}
\usepackage{pifont}% http://ctan.org/pkg/pifont
\usepackage{subfigure}
\usepackage{tikz}
\newcommand*\circled[1]{\tikz[baseline=(char.base)]{
            \node[shape=circle,draw,inner sep=0.5pt] (char) {\textmd{#1}};}}
\newcommand{\cmark}{\ding{51}}%
\newcommand{\xmark}{\ding{55}}%
\newcommand{\tb}[1]{\textbf{#1}}

\begin{document}

\maketitle
\begin{abstract}
The widespread popularity of equivariant networks underscores the significance of parameter efficient models and effective use of training data. At a time when robustness to unseen deformations is becoming increasingly important, we present H-NeXt, which bridges the gap between equivariance and invariance. H-NeXt is a parameter-efficient roto-translation invariant network that is trained without a single augmented image in the training set. Our network comprises three components: an equivariant backbone for learning roto-translation independent features, an invariant pooling layer for discarding roto-translation information, and a classification layer. H-NeXt outperforms the state of the art in classification on unaugmented training sets and augmented test sets of MNIST and CIFAR-10. 
\end{abstract}
%-------------------------------------------------------------------------

\section{Introduction}

\label{sec:intro}
\ac{cnn} \cite{lecun1998v2} are still one the most influential concepts in Computer Vision, and they are being actively researched \cite{liu2022v2, woo2023v2}. One of the key advantages that made \ac{cnn} dominant over fully connected networks is the weight sharing across the spatial dimension, i.e. the result of a translated input is an equally translated feature map. This property, called translation equivariance, makes \ac{cnn} implicitly robust and efficient with respect to translation. 
\par
This does not apply to other deformations, such as rotation, reflection, or affine transformation. Extensive data augmentation is a widely accepted practice to deal with these deformations \cite{krizhevsky2017v2}. As shown by Zeiler \etal \citep{zeiler2014v2}, \ac{cnn} filters trained on augmented data consist of rotated, scaled, and translated copies of one another. A question arises whether we can use parameters more effectively, as in the case of translation. Moreover, if we wanted to create a network robust to numerous deformations using data augmentation, the training set would grow exponentially and training then becomes prohibitively expensive.
\par
\ac{gcnn} formulated by Cohen and Weiler \cite{cohen2016v2} defined the notion of equivariance, which provides the basis for the effective use of weights. \ac{gcnn} led to the development of various methods that exploit symmetries by sharing weights, such as networks based on Steerable filters \cite{weiler2018v2,hoogeboom2018v2,bekkers2018v2,kondor2018v2,marcos2017v2} or Scattering transform \cite{sifre2012v2, sifre2013v2, bruna2013v2, sifre2014v2, oyallon2015v2}. Primarily for rotation, several applications have emerged in domains where objects naturally occur with different orientations, such as aerial photography \cite{han2021v2, ding2019v2, cheng2018v2, cheng2016v2}, microscopy imaging \cite{chidesterA2019v2, chidesterB2019v2}, or texture classification \cite{marcos2016v2}.
\par 
All of the aforementioned papers are based on benchmarks with small but augmented training sets. The achieved accuracy shows how effective the model is at using its weights on limited training data, but does not reflect the robustness to deformations that the network has not seen. The next natural step is to show model performance when it is trained only on the dataset without any augmentation. We call models trained without ever seeing an augmented image hard (mathematically) invariant networks, and they were first formulated by Khasonova and Frossard \cite{khasanova2017v2} and further improved by Hwang \etal \cite{hwang2021v2}.  
\par
This paper presents \ac{hnext}. \circled{1} Within the ablation study, we outline the most common problems encountered when working with unaugmented datasets and show solutions adopted by our model. \circled{2} \ac{hnext} surpasses the state of the art with an order of magnitude smaller number of parameters compared to its predecessor. \circled{3} Finally, our model is used to contextualize the invariant and equivariant roles within the rot-mnist dataset \citep{larochelle2007v2}. Source codes and datasets are available at \url{https://github.com/karellat/h-next}.

\section{Related Work}

\label{sec:related} 
Interest in transformation robust models has been growing in recent years, and these models were comprehensively summarized in a 2021 survey \cite{mumuni2021v2}. This article will focus on robustness to rotation and translation, as these deformations arise in a variety of natural tasks \cite{han2021v2, ding2019v2, cheng2018v2, cheng2016v2, chidesterA2019v2, chidesterB2019v2, marcos2016v2}. Articles addressing this topic can be neatly divided into three groups: soft (empirically) invariant, equivariant, and hard (mathematically) invariant. \par
Network (function) $\phi: X \rightarrow Y$ is equivariant with respect to the group $G_X$ iff
\begin{equation}
    \forall T \in G_X,\exists T' \in G_{\phi} \quad\quad \phi(T(x)) = T'(\phi(x))\,.
    \label{eq:equivariance}
\end{equation}
That is, transforming the input $x \in X$ by $T$ and then processing it by network $\phi$ is the same as first passing the same input $x$ through network $\phi$ and transforming the output by $T'$. Usually the equality of the groups $G_X$ and $G_{\phi}$ is considered in most cases, but $T$ and $T'$ can be different, $T=T'$ implying that the transformations commute with $\phi$.
In the case of invariant networks, $G_{\phi}$ is an identity. \footnote{For clarity, we will consider the transformation group to be the $360^{\circ}$roto-translation group $G_X = SE(2)$ and $G_X = G_{\phi}$ for equivariant networks.}

Soft invariance/equivariance refers to networks that do not satisfy the mathematical definition of \eqref{eq:equivariance}, but are forced to do so by common regularization techniques. This category includes the article by Lenc and Vivaldi \cite{lenc2015v2}, who minimize the distance between features of transformed and untransformed inputs in the loss function, Spatial Transformers \cite{jaderberg2015v2}, which estimate the deformation parameters of the input from input data and then normalize the input, or PDO-eCONvs \cite{shen2020v2} approximately equivariant convolution based on partial differential equations and many others \cite{gens2014v2, dieleman2016v2}.

Equivariant networks have become a broad topic since the publication of G-CNNs \citep{cohen2016v2}, which are equivariant to 90 degree rotations. Using steerable filters \cite{freeman1991v2} led to other discrete groups \cite{cohen2017v2,weiler2018v2,kondor2018v2,hoogeboom2018v2}, $360^{\circ}$ rotations \cite{worrall2017v2}, sphere surfaces \cite{cohen2019v2, esteves2018v2, cohen2018v2}, or 3D volumetric space \cite{deng2021v2}. E(2)-CNN \cite{weiler2019v2} then summarized the whole theory about steerable networks.  

Khasanova and Frossard \cite{khasanova2017v2} were the first to formalize networks with hard invariance properties by introducing the roto-translation invariant TigraNet, which consists of spectral convolutions and dynamic pooling. They propose a new type of datasets focusing on invariance properties, i.e. the training sets do not contain any augmented images, but the performance is measured on augmented test sets. Their experiments showed that neither equivariant nor soft invariant networks can produce roto-translation robust features in the unaugmented setting. In particular, they compared TigraNet with the classical \ac{cnn} \cite{boureau2010v2}, \ac{hnet} \cite{worrall2017v2}, Spatial Transformers \cite{jaderberg2015v2} and Deep Scattering Networks \cite{oyallon2015v2}. Their work was followed by the SWN-GCN architecture \cite{hwang2021v2} that uses an equivariant backbone corresponding to graph convolutional networks followed by \ac{gap} to obtain invariant descriptors. SWN-GCN outperformed TigraNet and became the state of the art on both MNIST and CIFAR-10. For both benchmarks, test sets were augmented by rotation, while only upright images were included in the training set.  
%
%Our architecture outperforms SWN-GCN \cite{hwang2021v2} on both benchmarks. We also demonstrate a more efficient use of network parameters. 

\section{Proposed Method}
\begin{figure}[h!]
    \centering
    \includegraphics[width=1.02\textwidth]{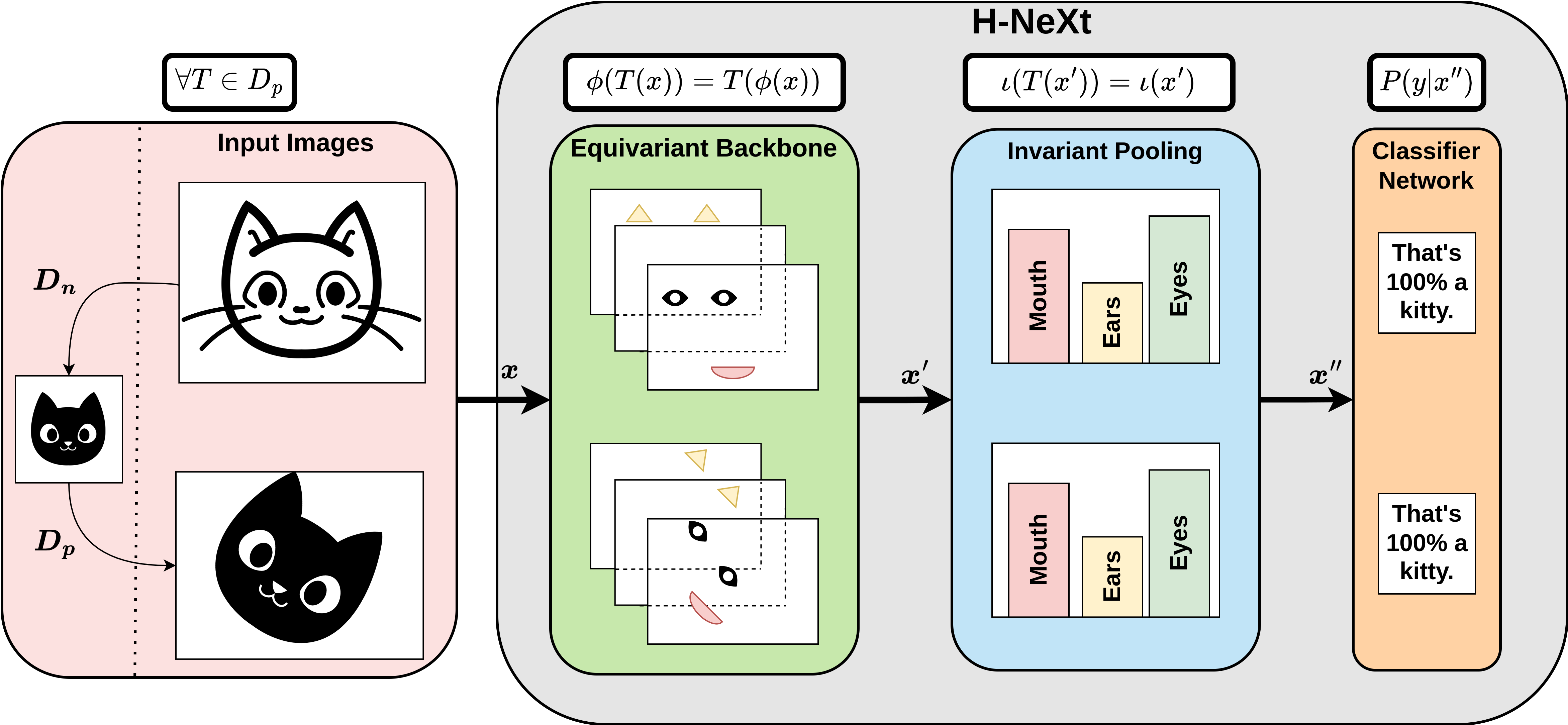}
    \caption{The proposed \ac{hnext} network invariant with respect to roto-translation ($D_p$) consisting of three parts: Roto-translation Equivariant Backbone, Roto-translation Invariant Pooling and Classifier Network.  }
    \label{fig:invnet}
\end{figure}

Our architecture is based on a general concept that includes \ac{hnext}, but also other models \cite{worrall2017v2, hwang2021v2, khasanova2017v2, singh2022v2, singh2023v2}. Suppose that objects within a single class can be described by a set of transformations decomposable into two subsets $D_p$ and $D_n$. While $D_p$ contains mathematically modelable transformations, $D_n$ contains all the others. For example consider the class of kittens, where $D_p$ could be the roto-translation group, and different fur, paw size, etc. would be transformations of $D_n$. The architecture, as illustrated in Figure \ref{fig:invnet}, consists of three parts: Equivariant Backbone, Invariant Pooling, and Classification Network. The backbone is equivariant with respect to $D_p$ $(=G_{X}=G_{\phi})$, and the goal is to create features that use parameters optimally by paying no attention to geometrically modelable variations, and leave that to the next layer, which is the invariant pooling with respect to $D_p$. The object class is predicted by a classifier network, which is typically an MLP. For \ac{hnext}, we always consider $D_p$ to be roto-translation and the backbone to be commutative with respect to $D_p$.

\subsection{Equivariant Backbone}
\label{sec:equivariant}
Traditional \ac{cnn} are translation equivariant, but not rotation equivariant, i.e. rotation changes not only the position of a feature, but also its value. To achieve rotation equivariance in our model, the values must be independent of input roto-translation.

\begin{figure}[h!]
    \centering
    \includegraphics[width=1.01\textwidth]{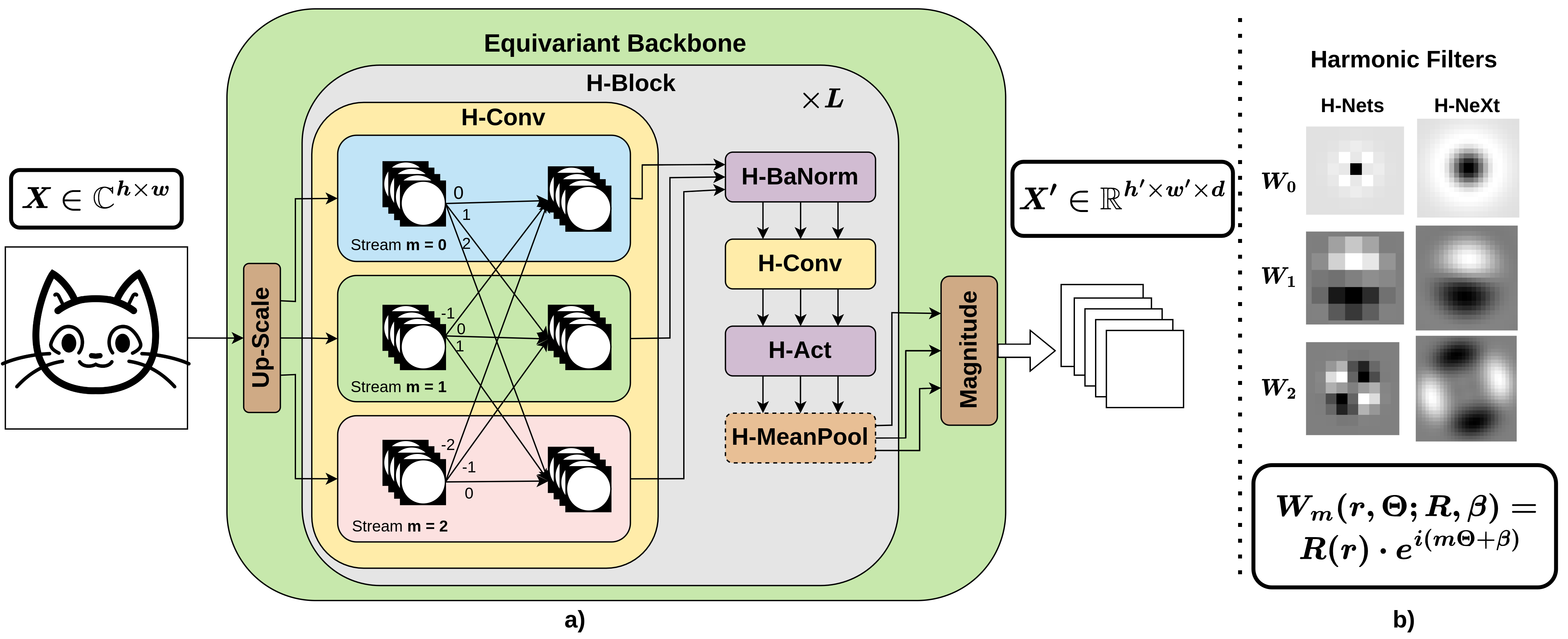}
    \caption{(a) Equivariant Backbone, the principal part of \ac{hnext} divided into its individual layers and (b), the comparison of \ac{hnet} and \ac{hnext} Harmonic Convolution Filters }
    \label{fig:backbone}
\end{figure}

The proposed backbone, shown in Figure \ref{fig:backbone}, is based on \ac{hnet} that operate on complex numbers\footnote{Note that complex polar form $z \in \mathbbm{C}; z=re^{i \varphi}$ ($r$ is the magnitude (distance) and $\varphi$ the angle (phase)) is used.}. Inside the backbone, the transformation $T_\theta \in D_p$, denoted by the rotation angle $\theta$ and an arbitrary translation, affects the channels (feature maps) $F^{T_\theta}_m$ with respect to the original channels $F^{0}_m$ as follows

\begin{equation}
    F^{T_\theta}_{m}(x,y) = e^{i m \theta} F^{0}_{m}(T_{\theta}(x,y)).
    \label{eq:rot_rank}
\end{equation}

Rotation changes only the phase, which is discarded at the end of the backbone, and the magnitude is used as the output. The channels are divided into streams according to $m \in \mathbbm{N}$, which is called rotation order.

%All of the \ac{hnext} blocks satisfy \eqref{eq:rot_rank} within three streams with $m = 0, 1, 2$.  TODO

\paragraph{Harmonic convolution (H-Conv)} H-Convs are convolution layers with filters limited to the circular harmonic family. Each filter is defined as
\begin{equation}
     W_{m_1} = W_{m_1} (r,\Theta; R, \beta) = R(r) \cdot e^{i(m_1 \Theta + \beta)},
     \label{eq:harm_filters}
\end{equation}
where $(r, \Theta) $ are polar coordinates, $R: \mathbbm{R^+} \rightarrow \mathbbm{R}$ is a radial function, $\beta \in [0, 2\pi)$ is the phase offset and $m_1 \in \mathbbm{Z}$ is the rotation order, representing axial filter symmetry as illustrated in Figure \ref{fig:backbone}b. Eq. \eqref{eq:harm_filters} implies that convolution of channels $F^{T_\theta}_{m_2}$ with $W_{m_1}$ follows 

\begin{equation}
    T_{\theta}(W_{m_1}(x,y) \star F^{T_\theta}_{m_2}(x,y)) = e^{i\theta (m_1 + m_2)} \cdot (W_{m_1}(T_{\theta}(x,y)) \star F^0_{m_2}(T_{\theta}(x,y)). 
    \footnote{Proof formulated by Worall \etal \cite{worrall2017v2} in Supplementary Material.}
    \label{eq:harm_conv}
\end{equation}

The result will then have the rotation order $m_1 + m_2$. Even though streams are otherwise processed separately in the backbone, Eq.~\eqref{eq:harm_conv} allows H-Convs to mix the streams without breaking \eqref{eq:rot_rank}. Resulting channels are reassembled into corresponding streams; note diagonal and horizontal connections inside H-Convs layer in Fig \ref{fig:backbone}a. To keep the number of streams fixed, H-Convs applies filters of rotation orders from $-m$ to $m$. 

Each further block of the backbone must satisfy \eqref{eq:rot_rank} to hold equivariance.

\paragraph*{Operating on magnitude (H-Act, H-BaNorm)}
The condition defined by \eqref{eq:rot_rank} is preserved if any arbitrary function, such as activation functions or batch normalization, is applied only to the magnitude, leaving the phase unchanged. \Eg Harmonic ReLU with bias $b$ for input $z$ is derived as follows 
\begin{equation}
      z = |z| \cdot  e^{\alpha i} \quad \mathit{HReLU_{b}}(z) = ReLU(|z| + b)  \cdot e^{\alpha i}.
\end{equation}

\paragraph{From continuous to discrete space (Up-Scale, H-MeanPool)}
All backbone blocks in continuous space satisfy the condition given by \eqref{eq:rot_rank}, but the images and network channels are discrete square grids, which leads to discretization inaccuracies. For example, Mean Pooling clearly follows \eqref{eq:rot_rank} in continuous space, but the rougher the feature discretization, the greater the difference for roto-translated input. To minimize this, we added an Up-Scale layer at the beginning of the backbone, which significantly reduces the inaccuracies of Mean Pooling and the other layers. 

\begin{figure}[h!]
    \centering
    \includegraphics[width=\textwidth]{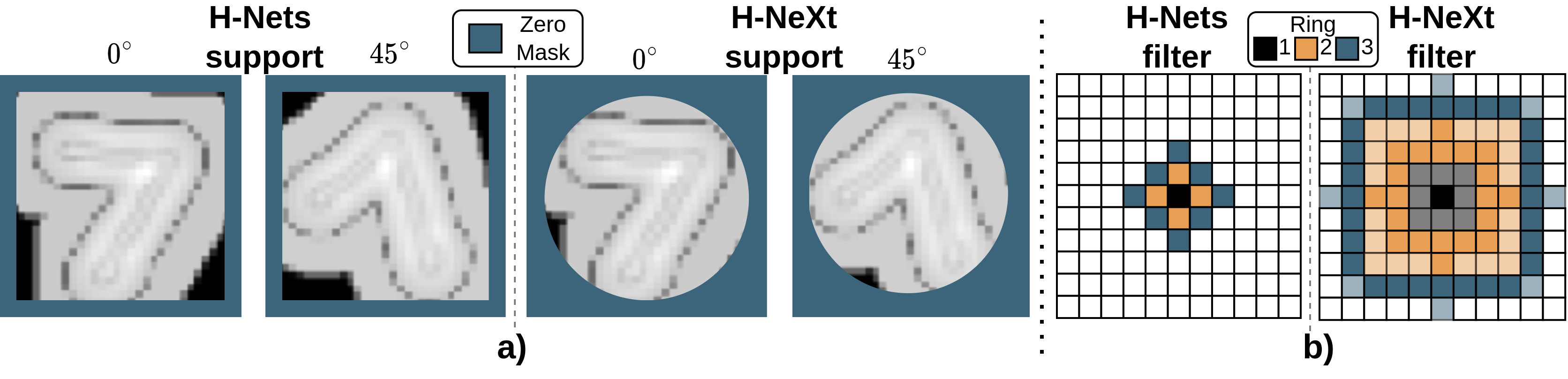}
    \caption{Comparison of \ac{hnet} and \ac{hnext} (a) channel shapes (b) convolution filters discretization}
    \label{fig:circ_prior}
\end{figure}

Not only the discrete features cause the equivariance violation, but also the filter discretization of the H-Convs has a non-negligible impact. As can be seen in Figure \ref{fig:backbone}b, the \ac{hnet} filters break the circular symmetry, which gets even worse with increasing rotation order. To avoid this, while keeping exactly the same number of weights, \ac{hnext} uses larger $(15\times15)$ filters. Each H-Conv filter has a learnable vector of size $n$ representing the radial function $R$. But in contrast with \ac{hnet}, where  $R$ could be seen as n rings with a distance of 1 px between each other, \ac{hnext} rings are evenly spread over the entire filter spatial size, as illustrated in Figures \ref{fig:backbone}b and \ref{fig:circ_prior}b. Using larger filters allows us to increase the maximum rotation order, contrary to Worall \etal's \cite{worrall2017v2} conclusions that streams with orders greater than 1 are not beneficial. 

These changes are referred to as \textbf{UP} in the experiments that follow in Section 
\ref{sec:experiments}. 

\paragraph*{Channel shape prior}
Another straightforward modification with a significant impact is to constrain the feature maps to a circular shape. The square channels used in most \ac{cnn} break the rotational equivariance due to the boundary effect, as shown in Figure \ref{fig:circ_prior}a. In \ac{hnext}, the constraint is implemented by applying a circular mask to each of the channels. These changes are referred to as \textbf{MASK} in the experiments that follow in Section \ref{sec:experiments}. 

\paragraph*{Backbone output} 
The last layer of the backbone discards the rotation-dependent phases, and returns only the magnitudes as roto-translation equivariant features. In \ac{hnet} the classification follows directly, which means that the pooling yields logits equal to the number of classes. Therefore, either only the $m=0$ stream is used, or the streams are merged by summing over the rotation orders, and in both cases aggregated by \ac{gap}.

Our experiments showed that for more complex tasks, we need to increase the complexity of the model, and simply adding more channels and blocks is not effective. Instead, we propose to use the magnitudes of each order separately, concatenate them, aggregate them by \ac{gap}, and feed them into a classification layer. The training loss decreases much faster due to a better gradient distribution across different rotation orders. These changes are referred to as \textbf{WIDE} in the experiments that follow in Section \ref{sec:experiments}.

\subsection{Invariant Pooling}
\begin{figure}[h!]
    \centering
    \includegraphics[width=\textwidth]{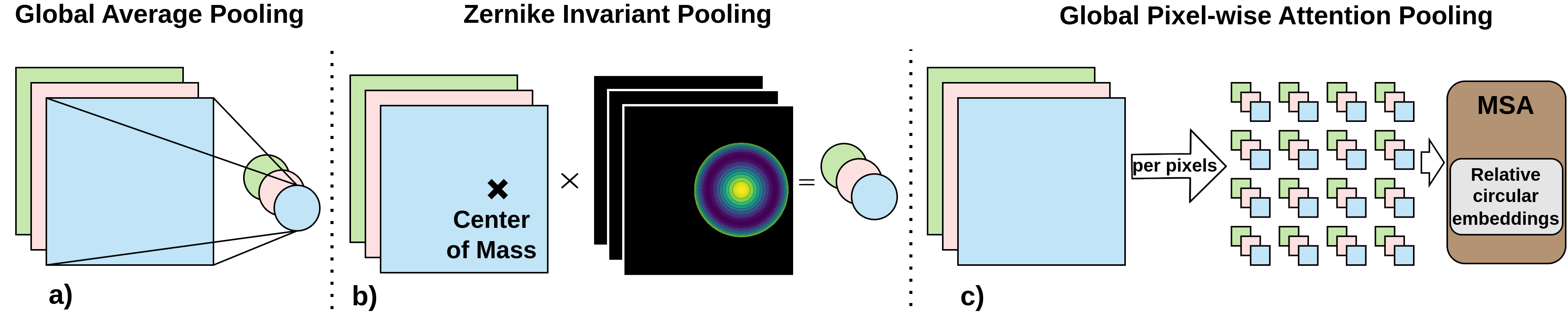}
    \caption{Three roto-translation invariant pooling options: (a) Global Average Pooling, (b) Zernike Invariant Pooling and (c) Invariant Multi-Head Self-Attention Pooling }
    \label{fig:poolings}
\end{figure}

We propose three approaches to obtain invariant features as illustrated in Figure \ref{fig:poolings}. The simplest approach is to use \ac{gap} as in \ac{hnet}, which is obviously roto-translation invariant. 

Zernike moment invariants \cite{teague1980v2}, as published by Singh  \etal \cite{singh2022v2, singh2023v2}, or other rotation moment-based invariants \cite{flusser2016v2, flusser2009v2, bedratyuk2022v2} can be used as pooling. Translation invariance is preserved by shifting the polynomials to the center of mass of each channel.

To further increase the complexity of the model, we propose to use \ac{msa} as an invariant pooling layer because it allows more intricate interactions. The input to \ac{msa} must be divided into patches, but roto-translation changes the pixel distribution into patches leading to invariance violation. To improve invariance, we work with a grid of the smallest possible windows $1 \times 1 \times d$, where $d$ is the number of input channels. For the same reason, we cannot add classical positional embeddings to the patches. To provide spatial information, relative circular embeddings similar to iRPE have been used \cite{wu2021v2}. The rest of the MSA is roto-translation invariant because the ordering of the patches doesn't matter. 

\section{Datasets}
\label{subsec:datasets}
The experiments were performed on three benchmarks summarized in Table \ref{tab:datasets}: \circled{1} Two newly generated datasets with unaugmented training set and randomly rotated test sets were created to demonstrate the invariance improvements. \circled{2} For comparison with the state-of-the-art, we follow the evaluation setup of SWN-GCN, which includes multiple test sets rotated by fixed angles ($0^\circ, 30^\circ, \dots, 330^\circ$). Unlike \circled{1}, the whole range of angles is not covered, \eg $45^\circ$, where accuracy is affected the most. \circled{3} To relate the equivariance and invariance tasks, \ac{hnext} was also tested on rot-mnist \cite{larochelle2007v2}. 

\begin{table}[h!]
    \centering
    \caption{Datasets used in experiments (Exp. id), including properties such as sizes, training (Rot. train) or test (Rot. test) set rotation, and whether the rotation performed (Rot. type) was at fixed (Fix) or at random angles (Rnd).}
    \label{tab:datasets}
    \bigskip
    \begin{tabular}{lrrrrrrrrr}
\hline
\thead{ \normalsize{Dataset} \\ \normalsize{name}}& \thead{\normalsize{Exp.} \\ \normalsize{id}} & \thead{\normalsize{Original} \\ \normalsize{dataset}} &  \thead{\normalsize{Train} \\ \normalsize{size}} &  \thead{\normalsize{Test} \\ \normalsize{size}}  &  \thead{\normalsize{Valid} \\ \normalsize{size}} & \thead{\normalsize{Rot. } \\ \normalsize{train} } & \thead{\normalsize{Rot.} \\ \normalsize{test}}  & \thead{\normalsize{Rot.} \\ \normalsize{type}} & Ref. \\

\hline\hline
mnist-rot-test& \circled{1} & MNIST & 50k & 10k & 10k &\xmark & \cmark & Rnd & Ours \\
swn-gcn-mnist & \circled{2}  & MNIST & 50k & 10k & 10k &\xmark & \cmark & Fix & \cite{hwang2021v2} \\
rot-mnist     & \circled{3}  & MNIST & 10k & 2k & 50k & \cmark & \cmark & Rnd & \cite{larochelle2007v2} \\
\hline
cifar-rot-test& \circled{1} & CIFAR10  & 42k & 10k & 8k &\xmark & \cmark & Rnd & Ours \\
swn-gcn-cifar & \circled{2} & CIFAR10 & 42k & 10k & 8k &\xmark & \cmark & Fix & \cite{hwang2021v2} \\
\hline
\end{tabular}

\end{table}

\section{Experiments}
\label{sec:experiments}
Three experiment settings were proposed in accordance with the datasets: \circled{1} The impact of \ac{hnext} architecture changes is investigated on our benchmarks. \circled{2} We present the state-of-the-art results in comparison with the SWN-GCN \cite{hwang2021v2} setup. \circled{3} The equivariance is tested on rot-mnist \cite{larochelle2007v2}, which is a standard benchmark for equivariance models.\footnote{Comprehensive experimental settings are listed in the Supplementary Material.}

\paragraph*{\circled{1} H-NeXt Ablation}
Models are trained on images in the upright position and evaluated on the randomly rotated test set. To evaluate the invariance capability of the network we propose to measure the difference between the validation accuracy at fixed angles $0^\circ$ (upright position) and at $45^\circ$. 

\begin{figure}[h!]
    \centering
    \includegraphics[width=1.03\textwidth]{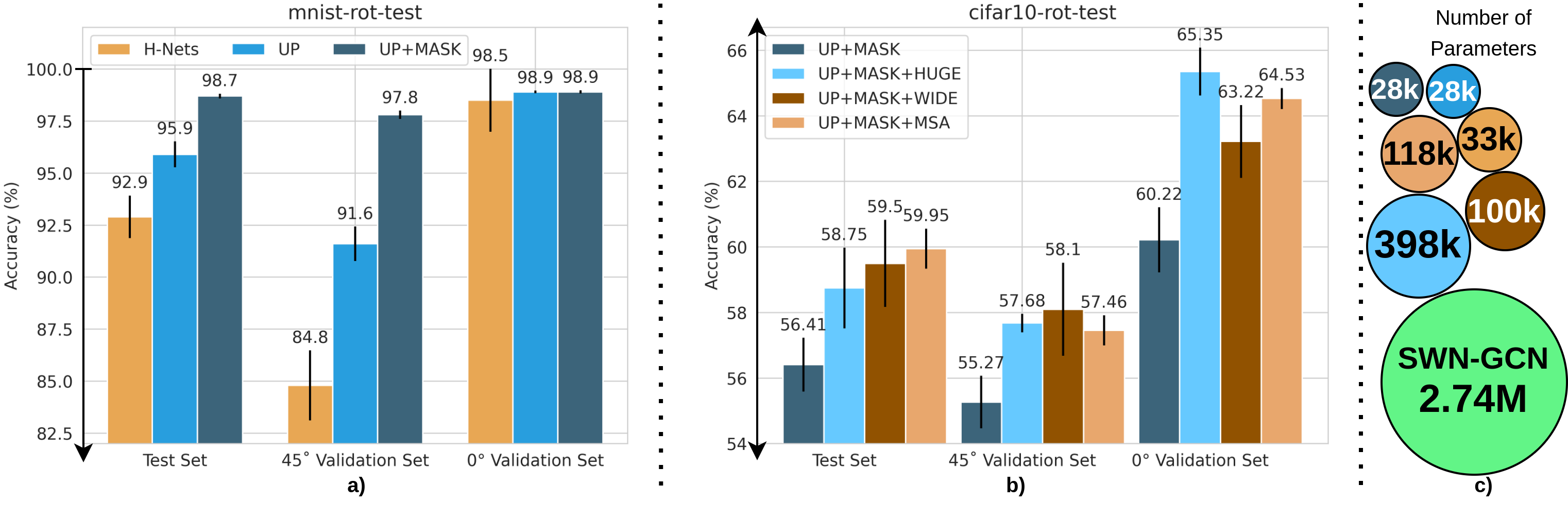}
    \caption{H-NeXt modifications impact on (a) mnist-rot-test accuracy (b) cifar10-rot-test accuracy and (c) model sizes. Test set contains randomly rotated images and validation sets are for specific angles. }
    \label{fig:rot-test-graph}
\end{figure}

The effects of each backbone enhancement on the mnist-rot-test are reported in Figure \ref{fig:rot-test-graph}. The performance is compared to \ac{hnet}, which has a significant gap of ${\sim}13\%$ between the classification of $0^\circ$  and $45^\circ$ rotated images. The \textbf{UP} model shrinks this gap to ${\sim}7\%$. Decreasing the discretization effects by including an Up-Scale layer and larger filters improves the invariance, as expected. Adding circular channels (\textbf{UP+MASK}) further reduces the gap to ${\sim}1\%$. Both of these changes lead to a significant improvement in test error and hence higher invariance. Further improvements on the mnist-rot-test may not be possible because of interpolation artifacts, as the model did not see any rotated data during training.

Different types of pooling are tested on cifar-rot-mnist\footnote{Note that CIFAR-10 images are circularly masked to remove rotation artifacts.}, the results are shown in Figure \ref{fig:rot-test-graph}b. \textbf{HUGE} has the worst performance on the test set ($58.75 \pm 1.23$), although it works best for $0^\circ$, it does not perform well for the other rotations. It is outperformed by \textbf{WIDE} ($59.50 \pm 1.33$) with four times fewer parameters, showing that increasing channels and blocks in \textbf{HUGE} is highly inefficient. By replacing the global average pooling with an invariant self-attention (\textbf{MSA}), the test accuracy is the highest ($59.95 \pm 0.61$). \textbf{MSA} increases performance for $0^\circ$, improves training stability and reduces variance, but slightly decreases $45^\circ$ accuracy. This shows that \textbf{WIDE} is less affected by interpolation, but on average \textbf{MSA} performs better.

\paragraph*{\circled{2} SWN-GCN comparison}
We train H-Next without any augmentations and measure the error on the test sets rotated by fixed angles $0^\circ, 30^\circ, 60^\circ, \dots$, following SWN-GCN evaluation setup \cite{hwang2021v2}.\footnote{A comprehensive accuracy list, including all angles, is included in the Supplementary Materials.} As can be seen in Table \ref{tab:sota-mnist}, \ac{hnext} for MNIST surpasses the current state-of-the-art  using just 28k parameters, which is a fraction of the 2.7M SWN-GCN parameters as illustrated in Figure \ref{fig:rot-test-graph}c.
In the case of CIFAR (see Table \ref{tab:sota-cifar}), the accuracy is improved by $7\%$  using models with up to 100k parameters, which is still significantly smaller than SWN-GCN.
Advanced pooling adds another $2\textrm{--}4\%$ using fewer parameters, where \textbf{\ac{msa}} works better for small rotations, but worse for large angles, and conversely the model \textbf{WIDE}, has more difficulties with the $0^\circ$ but works better at other angles.
\begin{table}[h]
    \centering
    \footnotesize
    \caption{Performance comparison on the MNIST invariance benchmark following the SWN-GCN \cite{hwang2021v2} setup. OA is the overall accuracy of all fixed angles. }
    \label{tab:sota-mnist}
    \bigskip
    \begin{tabular}{lrrrrrrrrrrr}
\hline
 MNIST Models & $0^\circ$ & $30^\circ$         & $120^\circ$ & $150^\circ$ & $210^\circ$ & $240^\circ$ & $300^\circ$ & $330^\circ$ & \tb{OA} \\ 
 \hline\hline
        RESNET-50 \cite{hwang2021v2}        & \tb{99.50} & 91.90      & 29.70 & 48.80 & 51.10 & 33.70 & 35.20 & 90.00 & 42.40 \\ 
        E(2)-CNN  \cite{weiler2019v2}       & 99.30      & 98.10      & 86.20 & 74.90 & 71.10 & 81.80 & 92.90 & 97.00 & 87.50 \\ 
        TIGRANET   \cite{khasanova2017v2}   & 89.10      & 82.70      & 82.70 & 79.80 & 82.70 & 79.80 & 82.70 & 79.80 & 85.10 \\
        SWN-GCN  \cite{hwang2021v2}         & 96.50      & 89.80      & 89.80 & 87.30 & 89.80 & 87.30 & 89.80 & 87.30 & 91.80 \\ 
        \ac{hnet} \cite{worrall2017v2}      & 98.70      & 89.41      & 89.41 & 90.55 & 89.41 & 90.55 & 89.40 & 90.55 & 92.89 \\ 
        \tb{UP+MASK}                        & 98.94      & \tb{98.55} & \tb{98.55} & \tb{98.55} & \tb{98.55} & \tb{98.55} & \tb{98.55} & \tb{98.54} & \tb{98.68} \\ 
\hline
\end{tabular}
\end{table}
\begin{table}[h]
    \centering
    \footnotesize
    \caption{Performance comparison on the CIFAR-10 invariance benchmark following the SWN-GCN \cite{hwang2021v2} setup. OA is the overall accuracy of all fixed angles} 
    \label{tab:sota-cifar} 
    \bigskip
    \begin{tabular}{lrrrrrrrrrrrrr}
\hline
      CIFAR Models  &    $0^\circ$ &   $30^\circ$ &  $120^\circ$ &  $150^\circ$ &  $210^\circ$ &  $240^\circ$ & $300^\circ$ &  $330^\circ$ &    \tb{OA}\\
\hline\hline
RESNET-50  \cite{hwang2021v2}  &\tb{85.10} & 54.50      & 27.50      & 26.90     & 27.00      & 24.90 & 33.20 & 52.50 & 36.10 \\
E(2)-CNN \cite{weiler2019v2}   & 77.10     & 57.80      & 34.40      & 30.80     & 31.90      & 35.40 & 45.00 & 56.00 & 46.20 \\
TIGRANET \cite{khasanova2017v2}& 38.90     & 37.00      & 37.00      & 36.80     & 37.00      & 36.80 & 37.00 & 36.80 & 38.10 \\
SWN-GCN  \cite{hwang2021v2}    & 51.30     & 49.60      & 49.60      & 50.10     & 49.60      & 50.10 & 49.60 & 50.10 & 50.50 \\
\tb{UP+MASK}               & 59.67     & 56.26      & 56.26      & 56.27     & 56.26      & 56.27 & 56.26 & 56.27 & 57.40 \\ 
\tb{UP+MASK+WIDE}          & 62.80     & \tb{60.31} & \tb{60.31} &\tb{60.35} & \tb{60.30} & \tb{60.36} & \tb{60.31} & \tb{60.36}  & 61.16 \\
\tb{UP+MASK+MSA}           & 64.15     & 60.09      & 60.08      & 60.13     & 60.09      & 60.13 & 60.09 & 60.13 & \tb{61.46} \\

\hline
\end{tabular}
\end{table}

\begin{figure}[h!]
    \centering
    \includegraphics[width=1.05\textwidth]{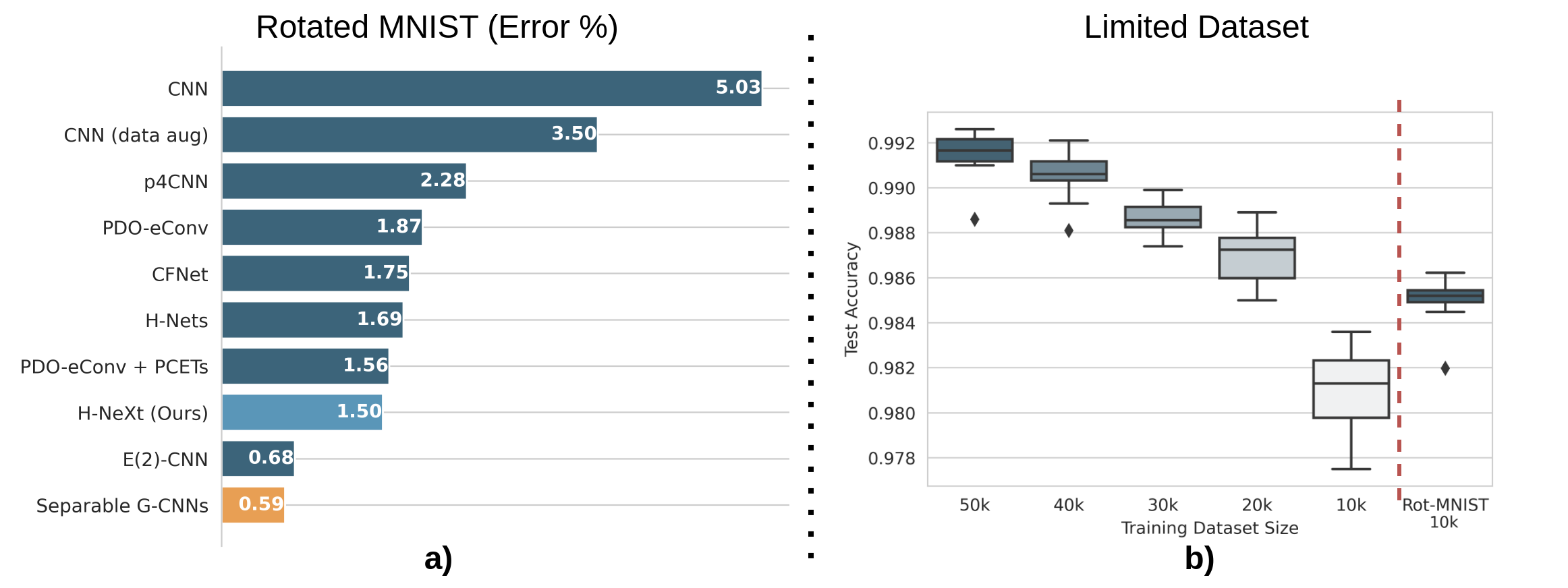}
    \caption{(a) Comparison of the models on the rot-mnist, namely: CNN \cite{cohen2016v2}, CNN(data aug) \cite{cohen2016v2}, G-CNN (p4CNN) \cite{cohen2016v2}, PDO-eConv \cite{shen2020v2}, CFNet \cite{chidesterA2019v2}, H-Nets \cite{worrall2017v2}, PDO-eConv+PCETs \cite{singh2023v2}, H-NeXt (Ours), E(2)-CNN\cite{weiler2018v2} and G-CNNs\cite{knigge2022v2}. (b) Comparison of \ac{hnext} performance between mnist-rot-test and rot-mnist (10k) on the limited training set.}
    \label{fig:rot-mnist}
\end{figure}

\paragraph*{\circled{3} Equivariant Context}  To establish a connection between the equivariance and invariance tasks, we include the results on rot-mnist in Figure \ref{fig:rot-mnist}a. Equivariant models (E(2)-CNN \cite{weiler2019v2}, \ac{hnet} \cite{worrall2017v2}) perform differently when trained on augmented and unaugmented data. Unlike hard invariant models (\ac{hnext}), the equivariant models leverage the information about rotation, when trained on augmented data.

For a comparison of mnist-rot-test and rot-mnist, we show the performance of \ac{hnext} on a different size with respect to the training size in Figure \ref{fig:rot-mnist}b. The accuracy of mnist-rot-test is lower than that of rot-mnist when the training size is the same (10k). Achieving invariance is more difficult when no rotated data is seen during training.

Moment invariant pooling (Fig. \ref{fig:poolings}) was tested in both settings \circled{1} and \circled{3} with the same performance as our model \textbf{UP+MASK}. This leads us to the conclusion that the circular support, which is used by both the Zernike Moment Pooling and the \textbf{UP+MASK} model, affects the model performance the most. 

\section{Conclusion}
\label{sec:conclusion} 
We propose \ac{hnext}, a roto-translation invariant network that achieves state-of-the-art results with fewer parameters, as validated by the invariance benchmarks according to the SWN-GCN \cite{hwang2021v2} setup. Our network targets tasks where a strong inductive bias is advantageous, for example, where roto-translated objects are naturally present. Restrictions on the architecture imposed by invariance constraints lower the maximum achievable model recognition capabilities. \ac{hnext} is thus not yet applicable in every context, but it is a step towards purely universal and invariant networks.

\paragraph*{Acknowledgements:} This work was supported by the Czech Science Foundation  under the grant No. GA21-03921S and by  the \emph{Praemium Academiae}. Special thanks to Adam Harmanec for proofreading and valuable comments. 

\bibliography{invariants_bmvc}
\begin{acronym}[TDMA]
\acro{cnn}[CNNs]{Convolutional Neural Networks}
\acro{cv}[CV]{Computer Vision}
\acro{hnet}[H-Nets]{Harmonic Networks}
\acro{hnext}[H-NeXt]{a roto-translation invariant model}
\acro{gcnn}[G-CNN]{Group Equivariant Convolutional Neural Networks}
\acro{gap}[GAP]{Global Average Pooling}
\acro{msa}[MSA]{Multi-Head Self-Attention}
\end{acronym}
\end{document}